\pdfoutput=1 
\documentclass{article}

\usepackage{arxiv}

\usepackage[utf8]{inputenc}
\usepackage[T1]{fontenc}
\usepackage{hyperref}
\usepackage[svgnames]{xcolor}
\hypersetup{
    colorlinks,
    linkcolor={DarkBlue},
    citecolor={DarkBlue},
    urlcolor={DarkBlue}
}
\usepackage{url}
\usepackage{booktabs}
\usepackage{amsfonts}
\usepackage{nicefrac}
\usepackage{microtype}
\usepackage{cleveref}
\usepackage{graphicx}
\usepackage[square,numbers]{natbib}
\usepackage{doi}

\usepackage{multirow}
\usepackage{xspace}
\usepackage{todonotes}
\usepackage{caption}
\usepackage{placeins}

\newcommand{\TFB}{FE\xspace}
\newcommand{\TFIT}{FiT\xspace}

\newcommand{\appref}[1]{App. \ref{#1}\xspace}

\title{Using Language Models on Low-end Hardware}

\author{Fabian~Ziegner \\
	Leipzig University \\
	\texttt{fz31dovo@studserv.uni-leipzig.de} \\
	\And
    Janos~Borst \\
	Leipzig University \\
	\texttt{janos.borst@uni-leipzig.de} \\
    \And
    Andreas~Niekler \\
	Leipzig University \\
	\texttt{andreas.niekler@uni-leipzig.de} \\
    \And
    Martin~Potthast \\
	Leipzig University and ScaDS.AI \\
	\texttt{martin.potthast@uni-leipzig.de} \\
}

\renewcommand{\shorttitle}{}

\begin{document}
\maketitle

\begin{abstract}
This paper evaluates the viability of using fixed language models for training text classification networks on low-end hardware. We combine language models with a CNN architecture and put together a comprehensive benchmark with 8~datasets covering single-label and multi-label classification of topic, sentiment, and genre. Our observations are distilled into a list of trade-offs, concluding that there are scenarios, where not fine-tuning a language model yields competitive effectiveness at faster training, requiring only a quarter of the memory compared to fine-tuning.
\end{abstract}


\section{Introduction}
\label{sect:introduction}

The transition to neural networks as primary machine learning paradigm in natural language processing~(NLP), and especially pre-training language models, became a major driver in NLP~tasks within the Digital Humanities. Many applications in fields ranging, among other things, from Library Science, Literature Studies or Cultural Studies have been dramatically improved and automation of text based tasks is becoming widely possible. Current state-of-the-art approaches utilize pre-trained neural language models, which are fine-tuned to a given set of target variables (i.e., by training all parameters of the language model). Training neural networks requires calculating a gradient for every layer and batch element, thus easily tripling the required memory. Those complex and multi-step architectures often use specific hardware, for example Graphics processing units (GPU), in order to be efficiently trained. This practice often exceeds the capabilities of end-user graphics cards. Large models easily require multiple GPUs called GPU clusters. This practice excludes various cultural institutions or universities that work in countries of the Global South or developing countries and have limited access to infrastructure or sufficient computing technology. Machine learning research originating in the Global North is frequently not as suitable for the south as its development occurred in a different environment \cite{Wall2021ArtificialII}. Finding out if there are nevertheless possibilities to run modern language models on older hardware is the motivation for writing this paper. Based on the fact that cheap and available end-user GPUs can still get by without fine-tuning, through fitting a fixed parameter language model this led us to the following questions:
(1)~Can older or cheaper graphics cards be used to train neural language models for text classification without fine-tuning?
(2)~How does their effectiveness compare to the state-of-the-art fine-tuning methods? 
(3)~What are the trade-offs between effectiveness and efficiency?
A less resource-intensive means to using language models is especially beneficial in cases where older graphics cards are available (e.g., due to the presently prohibitive upgrade costs), and where outsourcing to cloud services is not an option (e.g., for privacy, security, or budget reasons). Among others, our experiments show that the drop of performance for single-label topic classification is marginal at around~1\%; other tasks suffer more severe drops at around~10\%. At the lower end, training a fixed parameter language model is always superior to word embeddings.

\section{Related Work}
\label{sect:related_work}

Currently, text classification is mostly tackled using neural networks. Virtually every state-of-the-art result for text classification has been achieved using neural language models \cite{yang_xlnet_2019,aly_hierarchical_2019,Pal_2020}. Typically, a pre-trained language model is used, attaching an extra layer for projection to task-specific target variables, and fine-tuning both \cite{devlin-etal-2019-bert}. This either requires graphics cards with large memory, or low batch sizes. A substantial amount of research focuses on methods to downsize these pre-trained language models \cite{DBLP:journals/corr/abs-1910-01108,DBLP:journals/corr/abs-1301-3781,DBLP:conf/iclr/ClarkLLM20}. But, while these approaches reduce the number of parameters drastically, the memory during fine-tuning can still exceed 10~GiB due to gradient calculations.

Prior to language models, the most popular neural approach was to leverage pre-trained word embeddings, like GloVe \cite{pennington_glove:_2014} and Word2Vec \cite{DBLP:journals/corr/abs-1301-3781}, feed them to task-specific neural architectures \cite{xiao_label-specific_2019, kim_convolutional_2014}, and then train on the task data at hand. Although pre-trained word embeddings also range at around 100M~parameters, they are computationally efficient, since no gradient calculations are required. However, these approaches suffer from problems, such as out-of-vocabulary words and context insensitivity. We evaluate the performance of neural networks using a fixed-parameter language model as a middle ground between fine-tuning and word embeddings, investigating the trade-off between efficiency and effectiveness.

While we focus on either fine-tuning the language model or freezing it, there has been a number of works which deal with reducing the amount of parameters modified during fine-tuning while retaining performance. \cite{ben-zaken-etal-2022-bitfit,houlsby2019parameter,DBLP:journals/corr/abs-2110-04366}
\section{Experimental Setup}
\label{sect:experiments}

We employ the Huggingface library \cite{wolf-etal-2020-transformers} as a reference implementation for various state-of-the-art language models, and a PyTorch \cite{NEURIPS2019_9015} implementation of a CNN feature extraction module, which has proven useful for word-embedding-based models in different combinations \cite{kim_convolutional_2014, rios_few-shot_2018}. The following language models are compared to word embeddings: The base versions of BERT (\textit{bert-base-cased}) and RoBERTa (\textit{roberta-base}) with standard hyperparameters. Moreover, a heavily downsized BERT-Tiny, a 2-layer version of BERT-base (\textit{bert\_uncased\_L-2\_H-768\_A-12}), and a version with reduced hidden size and attention heads (\mbox{\textit{bert\_uncased\_L-12\_H-128\_A-2}}).

The CNN consists of $c$~convolutional layers with $k_i$~kernel sizes, $1\le i \le c$, and $f$~filters per layer. Its resulting feature vector is projected to the number of target classes of the corresponding dataset. Combinations of the CNN layer with either a language model or word embeddings are trained with and without fine-tuning. We use $c=4$ convolutional layers with kernel sizes of 3,4,5, and~6, and filter size~$f=100$. The CNN model is trained using Adam \cite{kingma_adam_2015} with a learning rate of~$5e-5$. The number of input tokens is set to~$200$. For multi-label tasks, a sigmoid activation of outputs and binary cross-entropy loss is used, and for single-label tasks, a softmax activation with categorical cross-entropy. For feature extraction, a batch size of~50 is used across all datasets while it has to be adjusted to~40 for fine-tuning to circumvent memory errors. We run each setting 3~times and report mean and standard deviation. Training epochs for each dataset are listed in Appendix~\ref{tab:epochs}.

The aforementioned models are evaluated on 8~datasets for a broad view of their effectiveness and efficiency compared to word embeddings. In both test cases, we feed the output of the language model into the CNN module and train both in combination. All datasets used are English. Each multi-label dataset has an unbalanced label distribution. The following datasets are included:

\textbf{AG News:}
News articles from the 4~largest topics of the corpus for a total of 30,000 training and 1,900 test samples per topic \cite{DBLP:conf/nips/ZhangZL15}.

\textbf{20NEWS:}
Messages from Usenet newsgroups with 20~topic classes. for a total of 11,314 training and 7,532 test samples \cite{lang_newsweeder_1995}.

\enlargethispage{-\baselineskip}
\textbf{DBpedia:}
An ontology dataset with 14~classes for a total 40,000 training and 5,000 test samples per class, randomly chosen from DBpedia~2014 \cite{DBLP:conf/nips/ZhangZL15}.

\textbf{TREC:}
Question classification dataset consisting of open-domain, fact-based questions. We use the versions with 6~and 50~classes, each containing 5,452 training and 500 test samples.

\textbf{Yelp:}
A sentiment classification dataset containing 650,000 Yelp reviews used as training samples, as well as 50,000 test samples. Each review may have a rating between~1 and 5~stars for classes.

\textbf{RCV1-v2:}
Topic classification dataset created by categorization of newswire stories. This version consists of 103~classes for a total of 23,149 training and 781,265 test samples with a label density of~3.12 \cite{lewis_rcv1_2004}.

\textbf{BlurbGenreCollection\_EN:}
Genre classification dataset using book blurbs made up of a short abstracts describing a given book with 152~classes for a total of 58,715 training and 18,394 test samples with a label density of~3.01 \cite{aly_hierarchical_2019}.

\textbf{Ohsumed:}
Medical dataset split into 23~different cardiovascular diseases for classes for a total of 7,643 training and 6,286 test samples with a label density of~1.64 \cite{Hersh94}.
\section{Results}

\textbf{Effectiveness:}
We report accuracy on the single-label datasets in Table~\ref{tab:single_performance_comparison} and micro Precision/Recall/F1 on multi-label datasets in Table~\ref{tab:multi_performance_comparison}. The results of feature extraction and fine-tuning are also compared to the current state-of-the-art results of the chosen datasets. On both single-label and multi-label datasets, fine-tuning on average performs better than feature extraction. In most cases, the smallest language model, BERT-Tiny, trained with feature extraction is on par with the word embeddings. However, when increasing model size or using fine-tuning, the word embeddings fall behind regarding recall. DBpedia is the only dataset on which BERT achieves better results with fine-tuning and feature extraction. We argue that this can be attributed to the much higher similarity between pre-training and downstream task compared to RoBERTa which was also observed by \cite{peters-etal-2019-tune}. While feature extraction can achieve good results on single-label data compared to fine-tuning it unfortunately falls short when training on multi-label data.

\begin{table}[h!]
\footnotesize
\centering
\renewcommand{\arraystretch}{1.0}
\renewcommand{\tabcolsep}{5pt}
\begin{tabular}{lcccccc}
    \toprule
    \bfseries{Method} & \bfseries{AG News} & \bfseries{20NEWS} & \bfseries{DBpedia} & \bfseries{TREC-6} & \bfseries{TREC-50} & \bfseries{YELP}\\
    \midrule
    GloVe-FE & 91.84 $\pm{0.18}$ & 79.85 $\pm{0.04}$ & 98.71 $\pm{0.01}$ & 92.33 $\pm{0.77}$ & 84.13 $\pm{0.57}$ & 58.75 $\pm{0.06}$ \\
    GloVe-FiT & 92.14 $\pm{0.19}$ & 80.06 $\pm{0.15}$ & 98.79 $\pm{0.02}$ & 92.33 $\pm{0.09}$ & 77.80 $\pm{0.71}$ & 59.63 $\pm{0.15}$ \\
    BERT-Tiny-FE & 90.74 $\pm{0.04}$ & 78.29 $\pm{0.28}$ & 98.74 $\pm{0.02}$ & 88.47 $\pm{0.19}$ & 71.13 $\pm{0.41}$ & 60.74 $\pm{0.09}$ \\
    BERT-Tiny-FiT & 92.92 $\pm{0.13}$ & 80.52 $\pm{0.19}$ & 99.01 $\pm{0.02}$ & 91.33 $\pm{0.25}$ & 81.33 $\pm{1.20}$ & 63.57 $\pm{0.09}$ \\
    BERT-L-2-FE & 93.03 $\pm{0.11}$ & 84.12 $\pm{0.26}$ & 99.18 $\pm{0.02}$ & 94.33 $\pm{0.38}$ & 78.40 $\pm{0.16}$ & 63.57 $\pm{0.06}$ \\
    BERT-L-2-FiT & 93.76 $\pm{0.15}$ & 84.90 $\pm{0.08}$ & 99.23 $\pm{0.02}$ & 95.00 $\pm{0.33}$ & 89.73 $\pm{0.25}$ & 65.13 $\pm{0.05}$ \\
    BERT-L-12-FE & 91.37 $\pm{0.09}$ & 78.88 $\pm{0.39}$ & 98.95 $\pm{0.01}$ & 90.27 $\pm{0.57}$ & 73.53 $\pm{0.47}$ & 60.77 $\pm{0.01}$ \\
    BERT-L-12-FiT & 93.47 $\pm{0.25}$ & 82.50 $\pm{0.47}$ & 99.08 $\pm{0.02}$ & 94.60 $\pm{0.43}$ & 87.93 $\pm{0.94}$ & 65.24 $\pm{0.18}$\\
    BERT-FE & 92.97 $\pm{0.06}$ & 84.25 $\pm{0.43}$ & 99.19 $\pm{0.02}$ & 94.87 $\pm{0.34}$ & 78.87 $\pm{0.62}$ & 63.25 $\pm{0.03}$ \\
    BERT-FiT & 94.01 $\pm{0.06}$ & 84.69 $\pm{0.29}$ & 99.26 $\pm{0.03}$ & 97.13 $\pm{0.25}$ & 92.00 $\pm{0.65}$ & 66.44 $\pm{0.17}$ \\
    RoBERTa-FE & 92.49 $\pm{0.06}$ & 85.81 $\pm{0.16}$ & 99.04 $\pm{0.01}$ & 78.53 $\pm{0.50}$ & 55.13 $\pm{0.62}$ & 64.16 $\pm{0.09}$\\
    RoBERTa-FiT & 94.84 $\pm{0.25}$ & 85.59 $\pm{0.3}$ & 99.21 $\pm{0.01}$ & 96.67 $\pm{0.09}$ & 90.67 $\pm{1.32}$ & 68.70 $\pm{0.05}$\\
    SOTA & \begin{tabular}{@{}c@{}}\textbf{95.55}  \\ \cite{yang_xlnet_2019} \end{tabular} 
         & \begin{tabular}{@{}c@{}}\textbf{93}     \\ \cite{wahba2023comparison} \end{tabular} 
         & \begin{tabular}{@{}c@{}}\textbf{99.38}  \\ \cite{yang_xlnet_2019} \end{tabular} 
         & \begin{tabular}{@{}c@{}}\textbf{98.07}  \\ \cite{cer-etal-2018-universal} \end{tabular} 
         & \begin{tabular}{@{}c@{}}\textbf{97.2}   \\ \cite{tayyar-madabushi-lee-2016-high}\end{tabular} 
         & \begin{tabular}{@{}c@{}}\textbf{73.28}  \\ \cite{DBLP:journals/corr/abs-1901-06610}\end{tabular} \\
    \bottomrule
\end{tabular}
\caption{Test Accuracy (\%) averaged over 3 runs on the single-label datasets. We compare feature extraction (FE) with our baselines and state-of-the-art (SOTA) models.}
\label{tab:single_performance_comparison}
\end{table}

\textbf{Memory:}
We train on an NVIDIA 1080Ti with 11GiB of VRAM. Furthermore, to measure the change when training on low-end hardware, we use an NVIDIA GTX 950 and an NVIDIA GTX 750 Ti with 2GiB of VRAM each. We compare the differences in memory usage by model between single-label and multi-label datasets using the feature extraction and fine-tuning approaches.
As shown in \appref{tab:memory_comparison}, when training with feature extraction the memory usage of the larger models is generally the same, hovering between 1.6-1.7 GiB. 
The same can be said about fine-tuning where the usage stays between 10.3-10.6 GiB when applying BERT and RoBERTa.
One must consider that even with a batch size reduction of 20\% these models take up almost all our available VRAM. Therefore, the actual memory savings with equalized batch sizes are larger.
While the memory usage of GloVe is the same as BERT-base concerning feature extraction, a large reduction is possible with the smallest BERT models for both feature extraction and fine-tuning. In the case of BERT-Tiny 1GiB of VRAM or less is needed in both cases.

\textbf{Time:}
There are two aspects to training time: Time per epoch and overall training time. 
To evaluate epoch time, we calculate the mean of each dataset's runs and divide by BERT-FE time to get relative values. The results are presented in \appref{tab:time_comparison}.
Generally speaking fine-tuning takes around \textbf{twice to thrice} as long per epoch than feature extraction.
On single-label datasets RoBERTa-FE overall takes longer to train than BERT, with the increase in training time conforming to the increase in model size. %
GloVe and the small BERT models only require a fraction of time compared to their larger counterparts. Time savings of around 95\% using feature extraction and 90\% using fine-tuning are possible with BERT-Tiny.
While per epoch time advantage is substantial for feature extraction, to be fair, we have to look at the net training time in \appref{tab:total_time_comparison}. 
To achieve the best results with feature extraction, it takes about 2-3 times more epochs to compete. This sometimes leads to feature extraction taking more overall time than fine-tuning, thus nullifying the gain per epoch.
When training on the low-end GPUs, the GTX 950 will take around 500\% the time to finish training, with the GTX 750 Ti taking even longer at around 700\%.

\medskip
\begin{table*}
\centering
\renewcommand{\arraystretch}{1.1}
\begin{tabular}{llccc}
    \toprule
    \bfseries{Dataset} &\bfseries{Method} & \bfseries{Precision} & \bfseries{Recall} & \bfseries{F1}\\
    \midrule
    \multirow{7}{*}{RCV1} & GloVe-FE & 90.33 $\pm{0.09}$ & 67.69 $\pm{0.13}$ & 77.38 $\pm{0.06}$\\
                          & GloVe-FiT & 90.70 $\pm{0.13}$ & 67.89 $\pm{0.10}$ & 77.65 $\pm{0.02}$\\
                          & BERT-Tiny-FE & 89.20 $\pm{0.12}$ & 70.21 $\pm{0.12}$ & 78.57 $\pm{0.05}$\\
                          & BERT-Tiny-FiT & 81.99  $\pm{0.45}$ & 78.22 $\pm{0.13}$ & 80.06 $\pm{0.16}$\\
                          & BERT-L-2-FE & 92.41 $\pm{0.02}$ & 73.37 $\pm{0.12}$ & 81.79 $\pm{0.07}$\\
                          & BERT-L-2-FiT & 84.99 $\pm{0.34}$ & 83.31 $\pm{0.31}$ & 84.14 $\pm{0.07}$\\
                          & BERT-L-12-FE & 91.58 $\pm{0.09}$ & 68.93 $\pm{0.14}$ & 78.66 $\pm{0.08}$\\
                          & BERT-L-12-FiT & 82.77 $\pm{0.17}$ & 82.79 $\pm{0.24}$ & 82.78 $\pm{0.06}$\\
                          & BERT-FE & 87.88 $\pm{0.18}$ & 77.97 $\pm{0.43}$ & 82.63 $\pm{0.16}$\\
                          & BERT-FiT & 86.12 $\pm{0.20}$ & 86.39 $\pm{0.19}$ & 86.26 $\pm{0.08}$\\
                          & RoBERTa-FE & 87.62 $\pm{0.11}$ & 81.34 $\pm{0.08}$ & 84.36 $\pm{0.04}$\\
                          & RoBERTa-FiT & 86.93 $\pm{0.52}$ & 87.30 $\pm{0.62}$ & 87.11 $\pm{0.07}$\\
                          & MAGNET \cite{Pal_2020} & - & - & \textbf{88.5}\\
    \midrule
    \multirow{7}{*}{Ohsumed} & GloVe-FE & 69.25 $\pm{0.43}$ & 56.99 $\pm{0.13}$ & 62.53 $\pm{0.19}$\\
                             & GloVe-FiT & 68.71 $\pm{0.22}$ & 59.14 $\pm{0.09}$ & 63.57 $\pm{0.14}$\\
                             & BERT-Tiny-FE & 65.79 $\pm{0.30}$ & 57.40 $\pm{0.34}$ & 61.31 $\pm{0.29}$\\
                             & BERT-Tiny-FiT & 60.97 $\pm{0.65}$ & 59.13 $\pm{0.81}$ & 60.03 $\pm{0.12}$\\
                             & BERT-L-2-FE & 74.46 $\pm{0.02}$ & 57.04 $\pm{0.24}$ & 64.60 $\pm{0.28}$\\
                             & BERT-L-2-FiT & 67.87 $\pm{0.81}$ & 65.23 $\pm{0.60}$ & 66.52 $\pm{0.21}$\\
                             & BERT-L-12-FE & 67.10 $\pm{0.04}$ & 56.02 $\pm{0.30}$ & 61.05 $\pm{0.17}$\\
                             & BERT-L-12-FiT & 66.08 $\pm{0.68}$ & 63.78 $\pm{1.08}$ & 64.89 $\pm{0.23}$\\
                             & BERT-FE & 71.25 $\pm{0.78}$ & 61.21 $\pm{0.07}$ & 65.84 $\pm{0.32}$\\
                             & BERT-FiT & 71.74 $\pm{0.80}$ & 69.75 $\pm{0.39}$ & 70.72 $\pm{0.31}$\\
                             & RoBERTa-FE & 69.67 $\pm{0.25}$ & 64.08 $\pm{0.26}$ & 66.76 $\pm{0.15}$\\
                             & RoBERTa-FiT & 73.78 $\pm{1.06}$ & 69.26 $\pm{2.15}$ & 71.74 $\pm{0.62}$\\
                             & BertGCN \cite{lin-etal-2021-bertgcn} & - & - & \textbf{72.8}\\
    \midrule
    \multirow{7}{*}{BGC\_EN} & GloVe-FE & 81.90 $\pm{0.34}$ & 50.98 $\pm{0.65}$ & 62.84 $\pm{0.40}$\\
                             & GloVe-FiT & 82.46 $\pm{0.19}$ & 53.49 $\pm{0.21}$ & 64.89 $\pm{0.12}$\\
                             & BERT-Tiny-FE & 80.64 $\pm{0.09}$ & 55.62 $\pm{0.28}$ & 65.83 $\pm{0.20}$\\
                             & BERT-Tiny-FiT & 71.50 $\pm{0.33}$ & 70.69 $\pm{0.08}$ & 71.10 $\pm{0.19}$\\
                             & BERT-L-2-FE & \textbf{85.31} $\pm{0.08}$ & 61.61 $\pm{0.02}$ & 71.55 $\pm{0.01}$\\
                             & BERT-L-2-FiT & 78.14 $\pm{1.29}$ & 74.56 $\pm{0.99}$ & 76.30 $\pm{0.11}$\\
                             & BERT-L-12-FE & 82.69 $\pm{0.12}$ & 54.89 $\pm{0.11}$ & 65.98 $\pm{0.12}$\\
                             & BERT-L-12-FiT & 72.75 $\pm{0.21}$ & 74.16 $\pm{0.20}$ & 73.44 $\pm{0.17}$\\
                             & BERT-FE & 76.87 $\pm{0.44}$ & 70.18 $\pm{0.27}$ & 73.37 $\pm{0.05}$\\
                             & BERT-FiT & 77.54 $\pm{0.23}$ & 78.64 $\pm{0.3}$ & 78.09 $\pm{0.17}$\\
                             & RoBERTa-FE & 73.72 $\pm{0.31}$ & 71.35 $\pm{0.34}$ & 72.52 $\pm{0.03}$\\
                             & RoBERTa-FiT & 78.41 $\pm{0.37}$ & \textbf{79.36} $\pm{0.32}$ & \textbf{78.88} $\pm{0.07}$\\          
                             & Caps. Network \cite{aly_hierarchical_2019} & 77.21 $\pm{0.54}$ & 71.73 $\pm{0.63}$ & 74.37 $\pm{0.35}$\\
    \bottomrule
\end{tabular}
\caption{Comparison between feature extraction (FE) and our baselines on multi-label datasets using Precision/Recall/F1 (\%). We include the best published result on the respective dataset for context.}
\label{tab:multi_performance_comparison}
\end{table*}
\FloatBarrier

\textbf{Trade-offs:} There are a number of observations to be made from these experiments, which are:

\emph{Feature extraction can be a viable option for single-label classification}, losing only $1.06\%$ of relative performance on average compared to its fine-tuning counterpart for topic classification, while using only 1.7 GiB VRAM. Sentiment and question classification seem to draw more benefit from fine-tuning with a relative performance gain of 7.08\% and 13\%, respectively.

\emph{Feature extraction loses significantly on multi-label classification}, $8\%$ decrease in performance on average, while also requiring more overall training time. The only argument for using feature extraction in multi-label tasks would be strong memory constrictions.

\emph{Feature extraction is a viable option if you have memory restrictions (even 2GiB or lower)}, since the larger feature extraction models hover around 1.65 GiB of VRAM with the smallest at 700MiB. For example, when having a restriction of 2GiB it is still better to use feature extraction on a large model than to fine-tune a small model.

\emph{Per epoch time efficiency}: In all feature extraction cases you get a better per epoch time efficiency. As discussed above, this amortizes when comparing the overall run times or hyperparameters. If time per epoch is a relevant metric (e.g., in split training sessions) this can be useful.

\emph{Very small language models are competitive}, when compared to word embeddings. The small memory footprint enables training on older hardware while maintaining a comparable performance combined with faster individual epochs.

No matter the hardware restrictions there is always a language model better and more efficient to use. If the language model crosses a minimum size it consistently outperforms word embeddings while maintaining a smaller memory footprint.
\section{Summary}
\label{sect:summary}
In this paper, we evaluated using language models without fine-tuning for text classification. We surveyed a set of publicly available datasets to distill a list of trade-offs regarding performance, time, and memory, to decide whether keeping a fixed language model is a viable option in scenarios with limited hardware resources.
We found that there are use cases in which using a larger model and not fine-tuning proves to be a good way to go, the main reason being memory restrictions.
The largest model in our experiments (RoBERTa) without fine-tuning and a batch size of 50 still has a memory footprint comparable to fixed word embeddings, while improving on every dataset by 6.61\% on average. Additionally, we hope that this paper will help researchers and cultural institutions that have limited access to technological resources to assess how they can still realize the tasks, perhaps with slightly lower quality. We also think that such studies will help to open the possibilities of Machine Learning to broader groups. Thus, we offer an alternative evaluation here that supports not only the best possible performance, but also efficient use of existing computational technology. In this way, 1.) older technology can be used and 2.) an assumed resource requirement can be reduced. We need to continue to promote such essays and work so that less wealthy countries and institutions can continue to have a stake in the practical application of machine learning.

\newpage
\bibliographystyle{unsrtnat}
\bibliography{anthology}

\appendix
\clearpage
\section{Appendix}
\label{sec:appendix}

\setcounter{table}{0}
\renewcommand{\thetable}{A\arabic{table}}

\renewcommand{\arraystretch}{1.0}

\begin{table}[h!]
\begin{minipage}[t]{.5\linewidth}
    \centering
    \footnotesize
    \begin{tabular}{lcc} 
        \toprule
        \bfseries{Dataset} & \bfseries{\TFB} & \bfseries{\TFIT} \\
        \midrule
        AGNews & 20 & 10 \\
        20NEWS & 300 & 100 \\
        DBpedia & 10 & 10 \\
        TREC-6 & 150 & 50 \\
        TREC-50 & 150 & 50 \\
        YELP & 15 & 5 \\
        RCV1 & 50 & 20 \\
        BGC\_EN & 40 & 20 \\
        Ohsumed & 300 & 80 \\
    \bottomrule
    \end{tabular}
    \caption{Number of epochs for every dataset.}
    \label{tab:epochs}
\end{minipage}\hfill
\begin{minipage}[t]{.5\linewidth}
    \centering
    \footnotesize
    \begin{tabular}{lcc} 
        \toprule
        \bfseries{Method} & \bfseries{\TFB} & \bfseries{\TFIT} \\
        \midrule
        GloVe & 1643 & 3941 \\
        BERT-Tiny & 693 & 1007 \\
        BERT-L-2 & 1371 & 3159 \\
        BERT-L-12 & 715 & 2419 \\
        BERT & 1667 & 10319 \\
        RoBERTa & 1729 & 10577\\
        \bottomrule
    \end{tabular}
    \caption{Average memory usage during training in MiB.}
    \label{tab:memory_comparison}
    \end{minipage}
\end{table}

\centering
\renewcommand{\arraystretch}{1.1}
\renewcommand{\tabcolsep}{6pt}
\begin{table}[h!]
\footnotesize
\begin{tabular}{lcccccccccccc}
    \toprule
    \bfseries{} & \multicolumn{2}{c}{BERT} & \multicolumn{2}{c}{RoBERTa} & \multicolumn{2}{c}{GloVe} & \multicolumn{2}{c}{BERT-Tiny} & \multicolumn{2}{c}{BERT-L-2} & \multicolumn{2}{c}{BERT-L-12}\\
    \bfseries{Dataset} &\bfseries{\TFB} & \bfseries{\TFIT} & \bfseries{\TFB} & \bfseries{\TFIT} & \bfseries{\TFB} & \bfseries{\TFIT} & \bfseries{\TFB} & \bfseries{\TFIT} & \bfseries{\TFB} & \bfseries{\TFIT} & \bfseries{\TFB} & \bfseries{\TFIT}\\
    \midrule
    AGNews & 1.0 & 2.62 & 1.25 & 2.59 & 0.04 & 0.24 & 0.05 & 0.11 & 0.21 & 0.52 & 0.12 & 0.37\\
    20NEWS & 1.0 & 1.96 & 1.08 & 1.74 & 0.03 & 0.15 & 0.05 & 0.08 & 0.15 & 0.3 & 0.13 & 0.27\\
    DBpedia & 1.0 & 2.58 & 1.24 & 2.54 & 0.04 & 0.23 & 0.05 & 0.1 & 0.21 & 0.51 & 0.12 & 0.36\\
    TREC-6 & 1.0 & 2.64 & 0.99 & 2.65 & 0.04 & 0.14 & 0.04 & 0.1 & 0.21 & 0.51 & 0.12 & 0.36\\
    TREC-50 & 1.0 & 2.63 & 1.0 & 2.64 & 0.04 & 0.14 & 0.04 & 0.1 & 0.21 & 0.51 & 0.12 & 0.36\\
    YELP & 1.0 & 2.67 & 0.99 & 2.69 & 0.04 & 0.15 & 0.05 & 0.1 & 0.23 & 0.54 & 0.13 & 0.38 \\
    RCV1 & 1.0 & 1.55 & 0.79 & 1.32 & 0.05 & 0.18 & 0.06 & 0.08 & 0.19 & 0.3 & 0.12 & 0.2\\
    Ohsumed & 1.0 & 1.92 & 0.8 & 1.74 & 0.05 & 0.27 & 0.06 & 0.09 & 0.19 & 0.38 & 0.12 & 0.28\\
    BGC\_EN & 1.0 & 2.11 & 1.08 & 1.97 & 0.05 & 0.27 & 0.06 & 0.1 & 0.2 & 0.42 & 0.12 & 0.29\\
    \bottomrule
\end{tabular}
\caption{Average time in seconds per epoch measured as multiples of BERT-FE.}
\label{tab:time_comparison}
\end{table}

\renewcommand{\arraystretch}{1.1}
\renewcommand{\tabcolsep}{5pt}
\begin{table}[h!]
\footnotesize
\begin{tabular}{lcccccccccccc}
    \toprule
    \bfseries{} & \multicolumn{2}{c}{BERT} & \multicolumn{2}{c}{RoBERTa} & \multicolumn{2}{c}{GloVe} & \multicolumn{2}{c}{BERT-Tiny} & \multicolumn{2}{c}{BERT-L-2} & \multicolumn{2}{c}{BERT-L-12}\\
    \bfseries{Dataset} &\bfseries{\TFB} & \bfseries{\TFIT} & \bfseries{\TFB} & \bfseries{\TFIT} & \bfseries{\TFB} & \bfseries{\TFIT} & \bfseries{\TFB} & \bfseries{\TFIT} & \bfseries{\TFB} & \bfseries{\TFIT} & \bfseries{\TFB} & \bfseries{\TFIT}\\
    \midrule
    AGNews & 4.32 & 5.62 & 5.43 & 5.58 & 0.17 & 0.51 & 0.78 & 0.90 & 3.65 & 4.51 & 2.08 & 3.17\\
    20NEWS & 10.08 & 6.86 & 11.33 & 6.11 & 0.34 & 0.50 & 1.93 & 0.85 & 5.46 & 3.32 & 3.58 & 2.42\\
    DBpedia & 10.51 & 26.74 & 12.91 & 26.43 & 0.41 & 2.39 & 2.04 & 3.35 & 8.76 & 16.02 & 5.07 & 11.20\\
    TREC-6 & 1.50 & 1.32 & 1.49 & 1.32 & 0.19 & 0.29 & 0.12 & 0.08 & 0.62 & 0.41 & 0.35 & 0.29\\
    TREC-50 & 1.50 & 1.32 & 1.50 & 1.32 & 0.19 & 0.29 & 0.12 & 0.08 & 0.62 & 0.41 & 0.35 & 0.29\\
    YELP & 17.65 & 15.74 & 17.51 & 15.82 & 3.61 & 6.10 & 4.30 & 4.02 & 7.99 & 6.39 & 7.07 & 6.72\\
    RCV1 & 6.31 & 3.92 & 4.93 & 3.36 & 0.34 & 0.46 & 1.16 & 1.48 & 3.56 & 5.31 & 2.32 & 3.70\\
    Ohsumed & 5.58 & 3.00 & 4.58 & 2.71 & 0.25 & 0.40 & 0.94 & 0.34 & 3.06 & 1.44 & 1.88 & 1.02\\
    BGC\_EN & 6.32 & 6.56 & 5.22 & 6.11 & 0.30 & 0.85 & 1.19 & 1.01 & 4.12 & 4.24 & 2.55 & 2.97\\
    \bottomrule
\end{tabular}
\caption{Average total training time per dataset in hours.}
\label{tab:total_time_comparison}
\end{table}

\end{document}